# Ontology Temporal Evolution for Multi-Entity Bayesian Networks under Exogenous and Endogenous Semantic Updating

Massimiliano Dal Mas

## Abstract

Increasingly sophisticated knowledge requires an inclusive knowledge representation that facilitates the integration of independently-generated information arising from different kind of efforts. The use of semantic web technologies addresses the reality of diverse interests on an ontology merging and supports knowledge discovery over that.

Significant research efforts in the Semantic Web are recently directed towards representing and reasoning with uncertainty and vagueness in ontologies.

It is a challenge for any Knowledge Base reasoning to manage ubiquitous uncertain ontology as well as updating with uncertain updating times, while achieving acceptable service levels at minimum computational cost. Uncertain updating times of ontologies and managing probabilistic uncertainty, possibilistic uncertainty, and vagueness in expressive description logics for the Semantic Web are well recognized to be one of the primary drivers of the costs in many ontology development.

This paper proposes an application-independent merging ontologies using the notions of temporal proposition, event, agent and role that can be used to specify any open interaction system towards merging different kind of ontologies.

A solution that uses Multi-Entity Bayesan Networks with SWRL rules, and a Java program is presented to dynamically monitor Exogenous and Endogenous temporal evolution on updating merging ontologies on a probabilistic framework for the Semantic Web.

If you have read this paper and wish to be included in a mailing list (or other means of communication) that I maintain on the subject, then send e-mail to: *me @ maxdalmas.com*

# 1 Introduction

Information merging is not only the most frequent application executed on the Web but it is also the basis of different types of applications. Considering collective intelligence of groups of individuals as a framework for evaluating and incorporating new experiences and information we often cannot merge such knowledge being tacit. Tacit knowledge underlies many competitive capabilities and is hard to articulate on discrete ontology structure. It is unstructured or unorganized, and therefore remains hidden.
Developing generic solutions that can merge the hidden knowledge from different type of ontologies is extremely complex. Moreover this will be a great challenge for the developers of semantic technologies.

This work aims to explore ways to make explicit and available the merging of knowledge hidden in the collective intelligence of different kinds of environment with different kind of ontologies.

The most important ingredient in any expert system is knowledge. The power of expert systems resides in the specific, high-quality knowledge they contain about task domains. Artificial Intelligence researchers will continue to explore and add to the current repertoire of knowledge representation and reasoning methods. But in knowledge resides the power. Because of the importance of knowledge in expert systems and because the current knowledge acquisition method is slow and tedious, much of the future of expert systems depends on breaking the knowledge acquisition bottleneck and in codifying and representing a large knowledge infrastructure.

This paper considers three major problems with both OWL 1 and OWL 2:

- **Uncertain Information**: not all of the information that needs to be represented on the Semantic Web is given in terms of definite statements. Any representation scheme intended to model real-world actions and processes must be able to cope with the effects of uncertain phenomena. Uncertainty is ubiquitous and it raises an unnecessary barrier to the development of new, powerful features for general knowledge applications.
- **Time**: No temporal operators is defined in OWL (with SWRL rules it's possible to have some partial temporal reasoning capabilities)
- **Open-World-Assumption**: OWL reasoning works under the open world assumption (everything we don't know is undefined). To deal with exogenous and endogenous semantic updating in merging different kind of ontologies it is necessary to simulate a closed world assumption (everything we don't know is false) by adding closure axioms before merging the ontologies.

The main contribution of this paper is to show the general framework for modelling and analyzing updating-time uncertainty, and the detailed analyses of updating system performance for an endogenous and an exogenous updating in a merging upper ontology

Another contribution of this work is a hybrid solution of the problem of monitoring the temporal evolution on merging different ontologies, based on a logic system that integrates First Order Logic (FOL) with Bayesian probability theory (MEBN) with a set of SWRL rules, and a Java program implemented using suitable OWL libraries (like the Jena Semantic Web Framework for Java9 or OWL API10).

Further research needs involve the use of analytical models to supplement the simulation studies reported here, as well as further studies of other stochastic updating-time regimes and mechanisms.

The paper is organized as follows:

- in the next section it is introduced the background on ontology and the Semantic Web languages used (OWL and SWRL);
- in Section 3 it is introduced the use of a Bayesian framework for probabilistic ontology to deal uncertainty in the Semantic Web;
- in Section 4 it is introduced a logic system that integrates Description Logic (DL) with Bayesian probability theory extending ordinary Bayesian networks (MEBN) to allow to represent knowledge as a collection of Bayesian network fragments (MFrag);
- in Section 5 it is presented an evolution of MEBN on a temporal knowledge base;
- in Section 6 it is presented the algorithm that was used to monitor the temporal evolution of a merging ontologies;
- in Section 7 it is presented the algorithm for the suggested Exogenous and Endogenous Semantic Updating;
- in Section 8 it is presented an actual example of a system specified using the defined meta-model;
- in Section 9 it is discussed the simulation results and analysis;
- then in Section 10 are introduced the related works;
- finally Section 11 draws some conclusions.

# 2 Background

**Ontologies** are a critical aspect of the Semantic, being a formal models of human domain knowledge they are difficult to build. An ontology is a formal specification of a conceptualization [1] that can be understood as an abstract representation of the world or domain we want to model for a certain purpose. Often there is no a discrete structure for single correct mapping on human knowledge.
Ontology-design tasks follows some rules but those cannot be comprehensive of the tacit knowledge.
Ontologies [2] are used to structure Knowledge in the Semantic Web.

**OWL** is the logical language recommended by the W3C for Semantic Web applications [3].
The OWL language consists of three sub-languages of increasing expressive power, namely OWL Lite, OWL DL and OWL Full. OWL Lite and OWL DL are, basically very expressive description logics; they are almost equivalent to the *SHIF(D+)* and *SHOIN(D+)* DLs. OWL Full is clearly undecidable because it does not impose restrictions on the use of transitive properties. Although the above DL languages are very expressive, they feature expressive limitations regarding their ability to represent vague and imprecise knowledge.

**OWL 2** is the latest standard in ontology languages from the World Wide Web Consortium (*www.w3.org/TR/owl2-profiles*) [4]. An OWL 2 ontology roughly corresponds to a *SROIQ(D)* knowledge base [5]. In the current paper we consider an extension of OWL with the Bayesan Networks (Pr-OWL) [6], which is a framework for covering vagueness. For the purposes of this paper, an ontology is regarded as a finite set of *SROIQ* axioms.

**SWRL** is a W3C recommendation for the rule language of the Semantic Web (*www.w3.org/Submission/SWRL*), which can be used to write rules to reason about OWL individuals and to infer new knowledge about those individuals. SWRL provides built-ins such as math, string and comparisons that can be used to specify extra contexts, which are not possible or very hard to achieve by OWL itself. SWRL is built on OWL DL and shares its formal semantics. In our practice, all variables in SWRL rules bind only to known individuals in an ontology in order to develop DL-Safe rules to make them decidable.

**Uncertain Information** is ubiquitous and is relate to any representation scheme that intended to model real-world actions and processes. Not all of the information that needs to be represented on the Semantic Web is given in terms of definite statements. Ontological information attached with statistical values can improve the behaviour of intelligent systems.

**Ontology Matching** allows to organize information from different ontologies for different sources. Semantic correspondences between the classes in the corresponding ontologies have to be determined and encoded in mappings to access information across these different sources.
Such mappings have been proposed by a number of approaches for automatically determining [7]. Machine learning techniques are used by mainly to compute the probability that two classes represent the same information.

# 3 Bayesian framework for probabilistic ontologies

A major shortcoming of existing Semantic Web technologies is their inability to represent and reason about uncertainty in a sound and principled manner.
To deal with uncertainty in the Semantic Web and its applications, many researchers have proposed extending OWL and the Description Logic (DL) formalisms with special mathematical frameworks. Researchers have proposed probabilistic [8, 9], possibilistic [10] and fuzzy extensions [11, 12, 13] among others.
A Bayesian framework for probabilistic ontologies provides a basis for plausible reasoning services in the Semantic Web. The Multi-Entity Bayesian Networks (MEBN) [6] is a logical formalism that combines first-order logic and probability as formal system that instantiates first-order Bayesian logic.
In MEBN Knowledge is expressed by a collection of MEBN fragments (MFrags) represents a conditional probability distribution of the instances of its resident random variables (RVs).
A collection of MFrags is organized into MEBN Theories (MTheories) ensuring the existence of a unique joint probability distribution over its random variables.
A probabilistic extension to the Web ontology language, Pr-OWL (pronounced as “prowl”), adds new definitions to current OWL while retaining backward compatibility with its base language.
Pr-OWL moves beyond deterministic classical logic [14, 15], having its formal semantics based on MEBN probabilistic logic [6]. By Pr-OWL is possible to model uncertainty in ontologies, it can capture imprecise and vague knowledge — for example, we can say that Milan is cold to a degree 0.7 rather than saying that Milan is either hot or not. Pr-OWL reasoning platform can capture and reason about such knowledge.

# 4 Description Logic developed using MEBN

A *knowledge base* contains both factual and heuristic knowledge:

- *Factual knowledge* represents the widely shared knowledge of the considered domain (as expressed in textbooks or journals, and commonly accepted in the particular field);
- *Heuristic knowledge* represent the individualistic judgmental knowledge of performance being less rigorous and more experiential (been the “good of practice” in the particular field).

The knowledge is formalized and organized by the *Knowledge representation*

Knowledge is almost always incomplete and uncertain. To deal with uncertain knowledge, a rule may have associated with it a *confidence factor* or a weight. The set of methods for using uncertain knowledge in combination with uncertain data in the reasoning process is called *reasoning with uncertainty*. An important subclass of methods for reasoning with uncertainty is called "fuzzy logic," and the systems that use them are known as "fuzzy systems." [11, 12, 13]
A knowledge base (KB) consists of a set of terminological axioms.
An ontology has to formally explicit a representation of a knowledge base (KB):

- Statistical regularities that characterize the domain;
- Inconclusive, ambiguous, incomplete, unreliable, and dissonant evidence related to entities of the domain;
- Uncertainty about all the above forms of knowledge.

A representation of a KB about a domain of application is formed by entities.
An entity is considered as any concept (real or fictitious, concrete or abstract) that can be described and reasoned about within the domain of application.
The KB meaning within computer science [16] is a model to describe the world in a set of:

- *Types* of entities in their domain;
- *Properties* of those entities;
- *Relationships* among entities with *processes and events* that happen with those entities.

We can consider the KB according to the Description Logic for an OWL ontology that consists of:

- a set of *types of entities* to describe types, which constitute the *Terminological Box*(*T-Box*);
- a set of *property axioms* to describe properties of individual entity, which constitute an *Assertion Box* (*A-Box*).
- and a collection of conditional formulas between roles to describe *relationships* among entities, which constitute a *RoleBox* (*R-Box*).

T-Box contains the intensional knowledge and is constituted by a set of axioms. All axioms are supposed to be a general concept inclusion axiom. T-Box statements tend to be more permanent within a knowledge base and tend to be stored in a data model or a metadata registry.
In contrast, the extensional knowledge is provided by a set of assertion that constitutes the A-Box [17]. Assertions provide information about “concrete objects” called entities, and can be concept membership assertions. A-Box statements are much more dynamic in nature and tend to be stored as instance data within transactional systems within databases.

The piece of knowledge represented by the production rule (R-Box) is relevant to the line of reasoning Expert systems whose knowledge is represented in rule form.
On this work was used SWRL (Semantic Web Rule Language) and Java code to overcome certain expressiveness limitations of OWL.

The DL framework used all along this paper is the usual one (see [66] for a detailed explanation). A knowledge base KB is defined as constituent by a T-Box, an A-Box, and an R-Box

$$KB = \sum \left\langle T\text{-}Box, A\text{-}Box, R\text{-}Box \right\rangle$$

An OWL ontology can be expressed using a MEBN logic where there are two kinds of domain-specific MFrags [18]:

- *generative MFrags* that corresponds to the *T-Box* (terminological box)
- *finding MFrags* that corresponds to the *A-Box* (assertional box)

The *T-Box*, considered as *generative domain-specific MFrags*, specify information about statistical regularities characterizing the class of situations to which an MTheory applies [6].
While the *A-Box*, considered as *finding MFrags*, can be used to specify particular information about a specific situation in the class defined by the generative theory (T-Box).
Findings can also be used to represent constraints assumed to hold in the domain [19, 20], although there are both computational and interpretation advantages to using generative MFrags when “constraint findings” can be avoided.
On the work some SWRL rules are introduced to deduce the truth value of temporal propositions on conditional formulas between roles describing *relationships* among entities on the R-Box.
By using SWRL rules with temporal built-ins is possible to bypass the lack on OWL temporal operators even if actually this does not provide full temporal reasoning capabilities

An *exogenous semantic* updating is a semantic change that comes from outside the model semantic and is unexplained by the inner semantic model. It involves an alteration of a semantic model that is autonomous by the semantic model itself.
While an *endogenous semantic* updating is explained within the model in which it appears.
The exogenous semantics approach [21] involves taking the semantic structures of a base logic (e.g. description logic) and combining them together (e.g. adding new structure) to provide the semantics for a higher-level ontology (e.g. for merging different kind of ontologies).
For example, in a simple ontology model of a dictionary: a new word is unknown by the inner ontology model and also leads to endogenous changes in demand that lead to create and changes the correlation between other ontology words.

The Semantic Web developers encodes all the information in an ontology filled with rules that say, essentially, that “Robert” and “Bob” are the same. But humans are constantly revising and extending their vocabularies, for instance at one times a tool might know that “Bob” is a nickname for “Robert”, but it might not know that some people named “Robert” use “Roby”, unless it is told explicitly. It is so necessary to have an exogenous semantic updating combined with a temporal evolution of the ontology. The A-Box needs to be reengineered with new assertions to model the updating of time, to obtain the closed-world assumption necessary for reasoning, and to model the actions achieved by the agents. When such updating is completed, a reasoner can be used to deduce that now Roby is a nickname of Robert.

# 5 Temporal evolution of the Ontology

An *upper ontology* (or foundation ontology) concern the more abstract concepts to practical use universal and domain independent, or independent single cognitive area (e.g. concepts such as "physical object", "event", "action" and so on, and properties as "position", "time of occurrence", "agent" and so on).
An agent can be defined as an autonomous entity, defined in terms of its behavior that acts to achieve defined goals according to the environment. (1)
An important aspect of the ontological merging system is the possibility to define if an agent should or should not perform, in a given interval of time, a merging on different ontologies.

Derived from the definition of Mfrag given in [22] we define an upper ontology considering a finite set of *Event (E)*, *Action (A)* and *Agent (N)* respect a fragment graph *G* and a set *D* of local distributions, one for each member of *Agent (N)* defined for the *TemporalEntity* of the Event.

$$F = (E, A, N, G, D)$$

The sets *E*, *A*, and *N* are pair wise disjoint (see Appendix for their definition according to Pr-OWL [6]).
The fragment graph *G* is an acyclic directed graph whose root nodes in *G* correspond (one-to-one) to random variables in *A U N*. Conditional probability distributions for the Agent (*N*) random variables is specified by local distributions [22].

The individual *TemporalEntity*, according to the W3C (www.w3.org/TR/owl-time), is introduced in the A-Box. The *TemporalEntity* is considered as a member of class *Event:Event(TemporalEntity)* [23]

```
TemporalEntity(T) <--> Instant(T) v Interval(T)
```

The above axioms can be expressed in OWL as:

```
<owl:Class rdf:ID="Instant">
	<rdfs:subClassOf rdf:resource="#TemporalEntity"/>
</owl:Class>

<owl:Class rdf:ID="Interval">
	<rdfs:subClassOf rdf:resource="#TemporalEntity"/>
</owl:Class>
<owl:Class rdf:ID="TemporalEntity">
	<owl:unionOf rdf:parseType="Collection">
		<owl:Class rdf:about="#Instant" />
		<owl:Class rdf:about="#Interval" />
	</owl:unionOf>
</owl:Class>
```

A-Box evolves in time, with the goal of *monitoring* the fulfillment or violation of obligations and prohibitions on merging ontologies.
A-Box is reengineered with new assertions to model the updating time and the action done by the agents allowing closed-world reasoning on classes to be merged on the upper ontology.

Assuming the closed-world assumption on reasoning we can consider that ontology has reached a stable state and the agents may perform queries to know what are their pending ontology that needs to be merged or prohibited or to react to their violation or fulfillment [23].
Some general classes of our ontology are used as domain or range of the properties used to describe temporal propositions and permanent changes; they are class *Event*, class *Action* and class *Agent*. In particular, an event may have as a property its *time* of occurrence. Some SWRL-rules can already be expressed in OWL 2, identifying all such Description Logic Rules leads to a decidable fragment.
In our case class *Action* is a subclass of *Event*, and has an own property to represent the *actor* of the action that can be decomposed in rules. Such properties maybe defined as follows:

$$Event \cap Agent \subseteq \bot \quad \Rightarrow \quad Event(x) \wedge Agent(x) \rightarrow$$
$$Action \subseteq Event \quad \Rightarrow \quad Action(x) \rightarrow Event(x)$$

Roles in *SROIQ* can be declared disjoint to state that elements related by one role must not be related by the other. To represent the updating time the individual *TemporalEntity* is introduced in the A-Box as a member of class *Event*: *Event(TemporalEntity)*.

# 6 Modeling time

Probabilistic Relational Models (PRM) provide a leap in representation power when compared with BNs, but PRMs alone cannot represent all that is needed for declarative representations that cover complex situations with tightly defined contexts (e.g. situations in which probability distributions are defined within very specific constrains). Therefore, to represent one of those specific situations in an ontology using PRMs as the underlying logic, either the instances of that ontology would also have to be declared (e.g. expressing that two items are not the same individual by referring to two actual instances of items123) or some combined approach would have to be used for constraining the context where the definitions apply (e.g. the use of Java program to model the updating time).
The need to declare all instances in advance makes the PRM solution unsuitable for most use cases for the Semantic Web, where the ontologies generally have only T-Box information (or occasionally a few built-in A-Box definitions) and the A-Box is left for each specific situation/application based on that ontology. Thus, the BNs solution seems to be a more appropriate way to build probabilistic ontologies.
Java program is used to model the time as a means to execute updating-time values under which the ontologies merging is update [23], to perform action at run-time by the interacting agents (monitoring the time evolution of the ontology merging), and to allow for closed-world reasoning on certain classes (see "Temporal evolution of the Ontology" Section for details).

---

(1)
Different definitions of *agents* are proposed in literature [24], these commonly include concepts such as:

- *persistence* (code runs continuously and decides for itself when some activity should be performed);
- *autonomy* (without human intervention agents have capabilities of decision-making - task selection, prioritization);
- *reactivity* (agents can react appropriately to the context in which they operate);
- *social ability* (agents may collaborate on a task with other agents)

Establishing such constraints is a vital asset for building probabilistic ontologies, since it allows one to express updating situations in which a given probability distribution holds. MEBN logic has a built-in form of representing such constraints (context nodes). Moreover MEBN makes it a simple and flexible technology that is also logically coherent (i.e. it can express a fully coherent joint probability distribution over instances that satisfy the constraints).

The main step followed by the Java program performs the following operations:

1. initiate the monitoring at time *t(0)* and close the extensions of the considered classes on which it is necessary to perform closed-world reasoning (the closed class is equivalent to the enumeration of all individuals that can be proved to be members of the class);
2. creating new individuals, cached in a queue, of the class *Event* in the A-Box, expressed by MFrags, for every event that happen in the merging between $t_i - 1$ and $t$
3. run a reasoner (as KAON2, Pellet 2.0 or HermiT to support for keys in OWL2) to deduce at the specified updating time instant all assertions that can be inferred (truth values of temporal propositions, states of merging, etc.);
4. update the merging relation between ontologies;
5. update the closure of the relevant classes;
6. increment the updating time *t* by 1 and go to the point 2.

# 7 Exogenous and Endogenous Semantic Updating

In this paper we consider regimes of sequential updating-times on a single-item ontology merging model managed by a continuously reviewed updating policy of different merged ontologies, but different updating-time regimes could be considered for further studies.

In a regime of sequential updating-times the updating of every ontology follow the queue determined by the temporal sequence of updates, which may be a reasonable assumption when a single ontology source is used.

Furthermore, the work considers the effects of exogenous versus endogenous ontology updating. An exogenous ontology updating is understood as a system in which the updating times are exogenously defined and unaffected by the flow of others ontologies updating in the merging system [25]. This may be interpreted as the case, where the single ontology is small compared to the upper ontology obtained by merging all the ontologies considered. So the single ontology updating only has a marginal effect on the capacity utilization of the upper ontology.

An endogenous ontology updating, on the other hand, is defined as a merging upper ontology in which the flow of ontology updating directly impacts the capacity utilization of the merged upper ontology. Moreover, the merged upper ontology in this case does have a significant effect on the updating times realized and vice versa. One simple way to model an endogenous system is to use a queuing mechanism.

We consider a single-item ontology system with the time updating under pure Poisson updating [26] controlled by a continuous review updating system. That kind of updating is stochastically dependent under the realistic assumption that it is time sequential. The service times (for endogenous updating ontology) and defined updating times (for exogenous updating ontology) is defined by a Gamma(1/μ, r) distribution [27].

The base-stock level on ontologies merged will be denoted by S. The base-stock policy places an updating merge whenever the updating position is below S corresponding to S-1 the updating point. As a standard setup for a base ontology merging model we assume a standard linear growing ontologies structure. Updating times are assumed to be sequential and stochastic. Sequential updating times imply that the updating ontology system is performed in the same sequence as they were placed.
The performance of systems respect to exogenous and endogenous merging ontologies, with the above specified characteristics, will be compared.
Demand generation and realization mechanisms on updating ontologies for the two systems are similar. However, design of ontology updating mechanisms for the two systems is obviously not similar.
An endogenous merging ontologies system can be modeled by using simple queuing mechanisms.
While for an exogenous merging ontologies system it is more complicated to model the sequential updating times.
As described in Zipkin [28], an exogenous system with sequential updating times at any time $t \geq 0$ defines (or exogenously generates) an updating time $U(t)$. To ensure that orders do not cross in time (e.g. that the ontological merging system is sequential) the updating time has to satisfy the condition that $t+U(t)$ is non-decreasing in $t$.

The main step followed by the Java program to obtain an exogenous sequential merging ontologies performs the following operations:

- if $t_n + U(t_n) \geq t_{n-1} + U(t_{n-1})$ a merging updating time is done precisely at $U(t_n)$
- if $t_n + U(t_n) \leq t_{n-1} + U(t_{n-1})$ a merging updating time is adjusted to $U(t_n) = (t_{n-1} + U(t_{n-1}))-t_n$

Like in many real-life situations an updating time to define for updating ontology received is decided regularly by the merging system. In the developed simulation model an updating time defined is decided upon once every day (during night, as usually done for update database).This updating time is initially drawn from a Gamma distribution, then it is checked for every merging request if its defined updating time satisfies the condition that is non-decreasing in $t$. If the condition is satisfied, then the defined updating time is used as it is. If the condition is not satisfied then the defined updating time for the current order is adjusted, such that the current updating is delivered at the same time as (or immediately after) the updating preceding it in the sequence of updating placed.

The main step followed by the Java program performs the following operations:

1. initiate the monitoring at time $t(0)$ and close the extensions of the considered Ontologies on which it is necessary to perform closed-world reasoning (the closed ontology is equivalent to the enumeration of all individuals that can be proved to be members of the ontology);
2. merging individuals, cached in a queue, of the class *Event* in the A-Box, according to the MTheories that express domain-specific ontologies that capture statistical regularities in a particular domain of application, for every event that happen in the merging between $t_j - 1$ and $t$;
3. run a Multi-Entity Decision Graphs (MEDG), a support for decision constructs in MEBN, to optimize a set of alternatives over the given constraints of a specific situation, to learn a Bayesian learning structure expressed as inference in MEBN theories;
4. update the merging relation between ontologies;

5. update the closure of the merged ontology;
6. increment the updating time *t* by 1 and go to the point 2.

Temporal propositions are used to represent the *content* and *condition* of ontology merging. [23]
They are a construct used to relate in two different ways a *proposition* to an *interval of time*. In the current OWL specification, we distinguish between positive and negative temporal proposition.
The classes necessary to model the updating time are *TemporalEntity*, with the two subclasses *TEPos* and *TENeg* used to distinguish between positive and negative temporal propositions.
*Positive temporal Entity* (a member of class *TEPos*) is used to represent a merging action, when the merging has to be performed within a given interval of time (with starting points $t_{start}$ and deadline $t_{end}$). If updating time satisfies the condition that is non-decreasing in *t*, then the defined updating time is used as it is. While *Negative temporal Entity* (a member of class *TENeg*) is used to model prohibitions on a merging action (when the merging must not be performed during a predefined interval of time).
If the condition is not satisfied then the defined updating time for the current merging is adjusted, such that it can occur immediately after the merging operation preceding it in the sequence.
Using SWRL rules is possible to determine the truth values of temporal propositions [23].
The ontology of the interaction system was created with the open source ontology editor Protege 4.1 beta. The editing of SWRL rules was created with Protege 3.4 and inserted their RDF/XML code in the ontology.

# 8 Example

Assume a situation where a user is looking for a Holyday on Sea based on two ontologies O1 and O2.
Let O1 be specified by the following axiom which specify that for each Event there is a keyword which is a subject. Furthermore, there is an Event about “Trip” which has the keyword Sea.

$$Event \subseteq \forall\ keyword.Subject$$
$$(Trip, Sea) : keyword$$
$$Trip : Event$$

Let O2 be specified by the following axioms which specify that Actions are even Events - Actions are contained in Event (see “Temporal evolution of the Ontology” Section) - and every concept in the knowledge base is about some topic. Furthermore we define a Trip as an Action and a Holyday as an Event.

$$Action \subseteq Event$$
$$\mathrm{T} \subseteq \forall About.Topic$$
$$Trip : Action$$
$$Holyday : Event$$
$$(Trip, Place)\, about$$
$$(Holyday, Duration)\, about$$

Without loss of generality, we can assume that O1 is the local ontology, i.e. the ontology being queried explicitly by the user, while O2 is an external ontology, i.e. the ontology of a travel agency. In order to integrate the information which is stored in both ontologies, mappings are needed. With a probabilistic matcher like InfoSleuth [29] or GLUE [30] mappings can be found which map the second ontology O2 to our local ontology O1.

$$
\begin{aligned}
&O1:Event(x) \leftarrow O2:Event(x)\ ;\ P(0.8)\\
&O1:Event(x) \leftarrow O2:Action(x)\ ;\ P(0.9)\\
&O1:Subject(x) \leftarrow O2:Topic(x)\ ;\ P(0.8)\\
&O1:keyword(x,y) \leftarrow O2:about(x,y)\ ;\ P(0.9)
\end{aligned}
$$

Taking for instance the first line of the above mapping: it basically says that all instances that are belonging to the concept Event in O2 are also belonging to the concept Event of O1 with the probability 0,8.
According to the Kolmogorov axioms of probability theory [26], the probability that instances belonging to *O2:Event* do not belong to *O1:Event* is 0,2. The probability that instances that do not belong to *O2:Event* belong to *O1:Event* needs to be derived by a matcher as well.
We have to consider that those probabilities can change so it should be possible to modify those according to the updating time of the ontologies.
If we pose a query, we want to get an answer that integrates the information of the local ontology O1 and the external ontology O2 in a defined time. So, in our example, to query the merging ontologies for all Events:

$$Event(x) \wedge keyword(x, Sea)$$

After having done the ontology merging, to get all relevant Event mentioned only in the external ontology O2, we have to consider:

- the Event about Trip with probability 1.0 because it is mentioned in the local ontology O1 and no mapping has been used for deriving it;
- the Event about Place in the external ontology O2 which was derived.

# 9 Simulation results and analysis

The system was modeled using a discrete event simulation. During the experiments we adjust 1/μ and r, on the Gamma distribution [27], to change the average updating time of the ontologies system and induce variability in merging times and the defined updating times. We run these experiments for a range of base-stock levels. For each experimental scenario, we observe effective updating-time characteristics and also study the effects on fill rate, average merged ontology and long-run average cost performance of the merged system, considering the processing time necessary for the merging.
So far the test done indicate that, in the presence of lost updating an exogenous merging system is more effective than an endogenous system, especially in case of long service times and large base-ontology levels. Also, for an exogenous sequential merging system was observed that average service time decreases with increasing ontology adjournment frequencies. In the future it could be of interest to model sequential ontology systems with other different systems. For such systems it could then be interesting to check for the service-time decrease observed here for high frequencies on ontology adjournment. Further research could analyze the characteristics of realized updating times under a wider range of experimental scenarios even involving analytical models for comparing the service-time regimes considered in this paper.

# 10 Related Works

The temporal evolution in merging uncertain information is an under explored area that arose from the ontology merging. In literature different kinds of approaches have been developed different aspects of the work. Two main classes are summarized below:

- *Upper Merged Ontology*

SUMO (Suggested Upper Merged Ontology) is the main famous free upper ontology intended as foundation ontology for a variety of computer information processing systems. It is organized for interoperability of automated reasoning engines and it is candidate for the "standard upper ontology" of IEEE working group 1600.1 (*www.ontologyportal.org*).
Other well known Upper Ontology: Cyc (*www.cyc.com*), WordNet (*wordnet.princeton.edu*), General Formal Ontology (*www.onto-med.de/en/theories/gfo*), Basic Formal Ontology (*www.ifomis.org/bfo*).
A series of a 3-month annual event on Ontology Community is helpful to be update on: *ontolog.cim3.net/cgi-bin/wiki.pl?OntologySummit2010.*

- *Ontology merging and alignment*

Ontology matching is a promising solution to the semantic heterogeneity problem. Different kind of ontology matching systems have been developed, many of those depend mainly on concept (dis)similarity measure derived from linguistic cues, multiple-word combinations, local and global meaning of a concept. An introduction of the topic is presented in [31] while a repository of information devoted to different aspects of ontology matching can be found online at *www.ontologymatching.org*

# 11 Conclusions

With the increased emphasis on semantic web use in heterogeneous systems, the study of ontology uncertainty is becoming important. This work has outlined a framework for studying this uncertainty when it is modelled managing probabilistic uncertainty on a merging of different kind of ontologies by means of updating-times. The work was then focused on a simulation study of a stochastic updating-time regimes of an upper ontology obtained by merging ontologies in a exogenous and endogenous way in which the updating-times is sequential and therefore dependent.
Updating-time for merging ontologies was driven using conditional obligations and prohibitions with starting points and deadlines monitored using OWL and SWRL.
A hybrid solution, based on OWL ontologies and MEBN was used for merging different ontologies driven by SWRL rules, and a Java program to monitor time evolution.
When studying the simulation results it was observed that variance of updating-times is very much dependent on the shape parameter of the Gamma distribution from which the defined updating-times was drawn. The long-run average cost (considering the processing time necessary for the merging) is not significantly affected by variance of updating-times in the endogenous merging system.
With an exogenous merging system, the variance of shorter updating-times does not have a significant influence on long-run average cost either.
However, that is not the case for longer updating-times and an exogenous merging system; the system performs worse under longer updating-times with higher variance than with lower variance.
Comparing the two merging systems, in general it appears that, in the presence of long updating-times, the performance of the exogenous ontology updating system is superior, particularly for high base-stock levels.
In the future it could be interesting to analyse the characteristics of realized updating-times under a wider range of different scenarios. Moreover, it could be interesting to model sequential updating systems with other mechanisms than the one considered in this study. For such systems it could be interesting to check for the updating-time decrease observed here for high updating frequencies. Further research could also involve analytical models for comparing the updating-time regimes considered in this paper.

**Massimiliano Dal Mas** is an engineer at the Web Services division of the Telecom Italia Group, Italy. His interests include: Ontology Engineering and Knowledge Base Systems, empirical computational linguistics and text data mining, user interfaces and visualization for information retrieval, automated Web interface evaluation. He received BA, MS degrees in Computer Science Engineering from the Politecnico di Milano, Italy. He won the thirteenth edition 2008 of the CEI Award for the best degree thesis with a dissertation on "Semantic technologies for industrial purposes" (Supervisor Prof. M. Colombetti).

# Appendix

- **Temporal entity**

```
TemporalEntity(T) <--> Instant(T) v Interval(T)
```

The above axioms can be expressed in OWL as:

```
<owl:Class rdf:ID="Instant">
	<rdfs:subClassOf rdf:resource="#TemporalEntity"/>
</owl:Class>


<owl:Class rdf:ID="Interval">
	<rdfs:subClassOf rdf:resource="#TemporalEntity"/>
</owl:Class>


<owl:Class rdf:ID="TemporalEntity">
	<owl:unionOf rdf:parseType="Collection">
		<owl:Class rdf:about="#Instant" />
		<owl:Class rdf:about="#Interval" />
	</owl:unionOf>
</owl:Class>
```

- **Event nodes**

```
    <owl:Class rdf:about="#Event">
        <rdfs:comment rdf:datatype="http://www.w3.org/2001/XMLSchema#string">EVENT nodes represent
constraints that must be satisfied by the physical object for the influences with the local
distributions of the fragment graph to apply.</rdfs:comment>
        <rdfs:subClassOf>
            <owl:Restriction>
                <owl:onProperty rdf:resource="#hasPossibleValues"/>
                <owl:allValuesFrom rdf:resource="#Entity"/>
            </owl:Restriction>
        </rdfs:subClassOf>
        <rdfs:subClassOf>
            <owl:Restriction>
                <owl:onProperty rdf:resource="#isContextInstanceOf"/>
                <owl:cardinality
rdf:datatype="http://www.w3.org/2001/XMLSchema#nonNegativeInteger">1</owl:cardinality>
            </owl:Restriction>
        </rdfs:subClassOf>
        <rdfs:subClassOf>
            <owl:Restriction>
                <owl:onProperty rdf:resource="#hasPossibleValues"/>
                <owl:someValuesFrom rdf:resource="#Entity"/>
            </owl:Restriction>
        </rdfs:subClassOf>
        <rdfs:subClassOf>
            <owl:Restriction>
                <owl:onProperty rdf:resource="#hasInnerTerm"/>
                <owl:allValuesFrom rdf:resource="#Event"/>
            </owl:Restriction>
        </rdfs:subClassOf>
        <rdfs:subClassOf>
            <owl:Class rdf:about="#Node"/>
        </rdfs:subClassOf>
        <rdfs:subClassOf>
            <owl:Restriction>
                <owl:onProperty rdf:resource="#isContextNodeIn"/>
                <owl:allValuesFrom rdf:resource="#Domain_MFrag"/>
            </owl:Restriction>
        </rdfs:subClassOf>
        <rdfs:subClassOf>
            <owl:Restriction>
                <owl:onProperty rdf:resource="#hasTemporalEntity"/>
                <owl:allValuesFrom rdf:resource="#Event"/>
            </owl:Restriction>
        </rdfs:subClassOf>
    </owl:Class>
```

- **Action nodes**

```
    <owl:Class rdf:about="#Action">
        <rdfs:comment rdf:datatype="http://www.w3.org/2001/XMLSchema#string">ACTION nodes delineate
the actions that agents should perform in a given interval of time been a copy of the Agent node
used as an input in a given MFrag. Each individual of class Action is linked with an individual of
class Agent via the property isActionInstanceOf.</rdfs:comment>
        <rdfs:subClassOf>
            <owl:Restriction>
                <owl:onProperty rdf:resource="#isParentOf"/>
                <owl:someValuesFrom rdf:resource="#Resident"/>
            </owl:Restriction>
        </rdfs:subClassOf>
        <rdfs:subClassOf>
            <owl:Restriction>
                <owl:onProperty rdf:resource="#hasArgument"/>
                <owl:allValuesFrom rdf:resource="#SimpleArgRelationship"/>
            </owl:Restriction>
        </rdfs:subClassOf>
        <rdfs:subClassOf>
            <owl:Restriction>
                <owl:onProperty rdf:resource="#isParentOf"/>
                <owl:allValuesFrom rdf:resource="#Resident"/>
            </owl:Restriction>
        </rdfs:subClassOf>
        <rdfs:subClassOf>
            <owl:Restriction>
                <owl:onProperty rdf:resource="#isParentOf"/>
                <owl:minCardinality
rdf:datatype="http://www.w3.org/2001/XMLSchema#nonNegativeInteger">1</owl:minCardinality>
            </owl:Restriction>
        </rdfs:subClassOf>
        <rdfs:subClassOf>
            <owl:Restriction>
                <owl:onProperty rdf:resource="#isActionInstanceOf"/>
                <owl:someValuesFrom>
                    <owl:Class>
                        <owl:unionOf rdf:parseType="Collection">
                            <owl:Class rdf:about="#BuiltInRV"/>
                            <owl:Class rdf:about="#Resident"/>
                        </owl:unionOf>
                    </owl:Class>
                </owl:someValuesFrom>
            </owl:Restriction>
        </rdfs:subClassOf>
        <rdfs:subClassOf>
            <owl:Restriction>
                <owl:onProperty rdf:resource="#hasInnerTerm"/>
                <owl:allValuesFrom rdf:resource="#Action"/>
            </owl:Restriction>
        </rdfs:subClassOf>
        <rdfs:subClassOf>
            <owl:Restriction>
                <owl:onProperty rdf:resource="#isActionNodeIn"/>
                <owl:allValuesFrom rdf:resource="#MFrag"/>
            </owl:Restriction>
        </rdfs:subClassOf>
        <rdfs:subClassOf>
            <owl:Restriction>
```

```
                <owl:onProperty rdf:resource="#isActionInstanceOf"/>
                <owl:cardinality
rdf:datatype="http://www.w3.org/2001/XMLSchema#nonNegativeInteger">1</owl:cardinality>
            </owl:Restriction>
        </rdfs:subClassOf>
        <rdfs:subClassOf>
            <owl:Class rdf:about="#Node"/>
        </rdfs:subClassOf>
        <rdfs:subClassOf>
            <owl:Restriction>
                <owl:onProperty rdf:resource="#isActionNodeIn"/>
                <owl:someValuesFrom rdf:resource="#MFrag"/>
            </owl:Restriction>
        </rdfs:subClassOf>
        <rdfs:subClassOf>
            <owl:Restriction>
                <owl:onProperty rdf:resource="#isActionInstanceOf"/>
                <owl:allValuesFrom>
                    <owl:Class>
                        <owl:unionOf rdf:parseType="Collection">
                            <owl:Class rdf:about="#BuiltInRV"/>
                            <owl:Class rdf:about="#Resident"/>
                        </owl:unionOf>
                    </owl:Class>
                </owl:allValuesFrom>
            </owl:Restriction>
        </rdfs:subClassOf>
    </owl:Class>
```

- **Agent nodes**

```
    <owl:Class rdf:about="#Agent">
        <rdfs:comment rdf:datatype="http://www.w3.org/2001/XMLSchema#string">AGENT nodes delineate the actions performed by the agents interacting within the system at run-time defined in the MFrag by their probability distributions.</rdfs:comment>
        <rdfs:subClassOf>
            <owl:Restriction>
                <owl:onProperty rdf:resource="#hasArgument"/>
                <owl:allValuesFrom rdf:resource="#SimpleArgRelationship"/>
            </owl:Restriction>
        </rdfs:subClassOf>
        <rdfs:subClassOf>
            <owl:Restriction>
                <owl:onProperty rdf:resource="#hasProbDist"/>
                <owl:someValuesFrom rdf:resource="#ProbDist"/>
            </owl:Restriction>
        </rdfs:subClassOf>
        <rdfs:subClassOf>
            <owl:Restriction>
                <owl:onProperty rdf:resource="#hasPossibleValues"/>
                <owl:allValuesFrom rdf:resource="#Entity"/>
            </owl:Restriction>
        </rdfs:subClassOf>
        <rdfs:subClassOf>
            <owl:Restriction>
                <owl:onProperty rdf:resource="#hasPossibleValues"/>
                <owl:someValuesFrom rdf:resource="#Entity"/>
            </owl:Restriction>
        </rdfs:subClassOf>
        <rdfs:subClassOf>
            <owl:Restriction>
                <owl:onProperty rdf:resource="#hasProbDist"/>
                <owl:allValuesFrom rdf:resource="#ProbDist"/>
            </owl:Restriction>
        </rdfs:subClassOf>
        <rdfs:subClassOf>
            <owl:Class rdf:about="#Node"/>
        </rdfs:subClassOf>
        <rdfs:subClassOf>
            <owl:Restriction>
                <owl:onProperty rdf:resource="#isAgentNodeIn"/>
                <owl:allValuesFrom rdf:resource="#MFrag"/>
            </owl:Restriction>
        </rdfs:subClassOf>
        <rdfs:subClassOf>
            <owl:Restriction>
                <owl:onProperty rdf:resource="#hasInnerTerm"/>
                <owl:allValuesFrom rdf:resource="#Resident"/>
            </owl:Restriction>
        </rdfs:subClassOf>
        <rdfs:subClassOf>
            <owl:Restriction>
                <owl:onProperty rdf:resource="#isArgTermIn"/>
                <owl:allValuesFrom rdf:resource="#ArgRelationship"/>
            </owl:Restriction>
        </rdfs:subClassOf>
        <rdfs:subClassOf>
            <owl:Restriction>
```

```
                <owl:onProperty rdf:resource="#isParentOf"/>
                <owl:allValuesFrom rdf:resource="#Resident"/>
            </owl:Restriction>
        </rdfs:subClassOf>
        <rdfs:subClassOf>
            <owl:Restriction>
                <owl:onProperty rdf:resource="#isAgentNodeIn"/>
                <owl:minCardinality
rdf:datatype="http://www.w3.org/2001/XMLSchema#nonNegativeInteger">1</owl:minCardinality>
            </owl:Restriction>
        </rdfs:subClassOf>
        <rdfs:subClassOf>
            <owl:Restriction>
                <owl:onProperty rdf:resource="#hasParent"/>
                <owl:allValuesFrom>
                    <owl:Class>
                        <owl:unionOf rdf:parseType="Collection">
                            <owl:Class rdf:about="#Input"/>
                            <owl:Class rdf:about="#Agent"/>
                        </owl:unionOf>
                    </owl:Class>
                </owl:allValuesFrom>
            </owl:Restriction>
        </rdfs:subClassOf>
        <rdfs:subClassOf>
            <owl:Restriction>
                <owl:onProperty rdf:resource="#isAgentNodeIn"/>
                <owl:someValuesFrom rdf:resource="#MFrag"/>
            </owl:Restriction>
        </rdfs:subClassOf>
    </owl:Class>
```